\documentclass[sigconf]{acmart}

\AtBeginDocument{%
  \providecommand\BibTeX{{%
    \normalfont B\kern-0.5em{\scshape i\kern-0.25em b}\kern-0.8em\TeX}}}

\copyrightyear{2021}
\acmYear{2021}
\setcopyright{acmcopyright}\acmConference[MM '21]{Proceedings of the 29th ACM International Conference on Multimedia}{October 20--24, 2021}{Virtual Event, China}
\acmBooktitle{Proceedings of the 29th ACM International Conference on Multimedia (MM '21), October 20--24, 2021, Virtual Event, China}
\acmPrice{15.00}
\acmDOI{10.1145/3474085.3475651}
\acmISBN{978-1-4503-8651-7/21/10}

\acmSubmissionID{mfp2637}

\usepackage{bm}
\usepackage{multirow}
\usepackage{graphics}
\usepackage{subfigure}
\usepackage{gensymb}
\usepackage{balance}

\begin{document}
\fancyhead{}

\title{Progressive and Selective Fusion Network\\
for High Dynamic Range Imaging}

\author{Qian Ye}
\affiliation{%
  \institution{Graduate School of Information Sciences,\\ Tohoku University}
 \city{Sendai}
  \country{Japan}
}
\email{qian@vision.is.tohoku.ac.jp}

\author{Jun Xiao}
\affiliation{%
  \institution{Department of Electronic and Information Engineering, The Hong Kong Polytechnic University}
  \streetaddress{}
  \city{Hong Kong}
  \state{}
  \country{China}
  \postcode{}
}
\email{jun.xiao@connect.polyu.hk}
\author{Kin-man Lam}
\affiliation{%
  \institution{Department of Electronic and Information Engineering, The Hong Kong Polytechnic University}
  \streetaddress{}
  \city{Hong Kong}
  \country{China}
}
\email{enkmlam@polyu.edu.hk} 

\author{Takayuki Okatani}
\authornote{The corresponding author.}
\affiliation{%
  \institution{Graduate School of Information Sciences,\\ Tohoku University / RIKEN Center for AIP}
  \city{Sendai}
  \country{Japan}
}
\email{okatani@vision.is.tohoku.ac.jp}


\begin{abstract}
This paper considers the problem of generating an HDR image of a scene from its LDR images. Recent studies employ deep learning and solve the problem in an end-to-end fashion, leading to significant performance improvements. However, it is still hard to generate a good quality image from LDR images of a dynamic scene captured by a hand-held camera, e.g., occlusion due to the large motion of foreground objects, causing ghosting artifacts. The key to success relies on how well we can fuse the input images in their feature space, where we wish to remove the factors leading to low-quality image generation while performing the fundamental computations for HDR image generation, e.g., selecting the best-exposed image/region. We propose a novel method that can better fuse the features based on two ideas. One is multi-step feature fusion; our network gradually fuses the features in a stack of blocks having the same structure. The other is the design of the component block that effectively performs two operations essential to the problem, i.e., comparing and selecting appropriate images/regions. Experimental results show that the proposed method outperforms the previous state-of-the-art methods on the standard benchmark tests. 
\end{abstract}

\ccsdesc[500]{Computing methodologies~Image processing}
\ccsdesc[300]{Computing methodologies~Computational photography}
\keywords{Image processing; High dynamic range imaging}

\maketitle

\section{Introduction}
As real-world scenes usually have a range of luminosity beyond the dynamic range of imaging devices, standard digital cameras can only produce low dynamic range (LDR) images containing under-exposure and over-exposure regions where the detailed information is missing.
There are demands for high dynamic range (HDR) imaging from various fields
such as movie \cite{rempel2009video} and computer rendering \cite{anderson2007critters}, etc. 
There are special cameras that can capture HDR images, which tend to be
expensive. 
Thus, there are methods for generating an HDR image from a series of LDR images captured by a standard camera with different exposure settings.
While they can produce high-quality HDR images for static scenes, these methods tend to yield images with many ghosting artifacts for dynamic scenes or even for static scenes when the input images are captured by a hand-held camera.
Numerous efforts have been made so far to remove the ghosting artifacts in the HDR image reconstruction. There are several methods that attempt to detect motion regions in the input LDR images and then remove these regions \cite{khan2006ghost,heo2010ghost,yan2017high}. However, they tend to work well only when the motion in the input images is relatively small. When there is large motion, a large number of image pixels need to be removed, which results in incorrect reconstruction due to missing information about these pixels. 
There are also methods that align the input LDR images using the optical flows to a reference image from the others before merging them \cite{bogoni2000extending, hu2013hdr,hafner2014simultaneous}. While these methods can handle larger image motion, their performance highly depends on the accuracy of the estimated optical flows. When the motion regions are either over or under-exposed, there tend to emerge noticeable artifacts in the resulting HDR images. 

More recent studies employ convolutional neural networks (CNNs) and formulate the problem in an end-to-end fashion. They train CNNs to learn the direct mapping from multiple LDR images to an HDR image using appropriate training data \cite{wu2018deep,yan2019attention, yan2020deep}. Although they achieve better performance than the above methods, they still suffer from the ghosting artifacts when there is large motion in the input LDR images.  

We can think of the current CNN-based methods employing the same approach to the problem, i.e., fusing the input images in their feature space and then reconstructing an HDR image from the fused feature, typically using an encoder-decoder network. It relies on the feature fusion to solve the two fundamental problems, i.e.,  selecting well-exposed images/regions from the input images and correcting/eliminating the misalignment of the images plus possible occlusions due to object/image motion. Previous studies perform this feature fusion in a single step using a relatively simple method such as summation and concatenation. We think this leads to suboptimal feature fusion, causing ghosting artifacts.  

In this study, we propose a novel network named {\em progressive and selective feature network} (PSFNet) to resolve the above issue. PSFNet employs i) a multi-step approach that progressively fuses features and ii) a more suitable mechanism for feature fusion. 

For (i), we split the difficult problem of feature fusion into multiple steps, by which we intend to make it easier for the network to learn to solve it. It is analogous to nonlinear optimization algorithms that update parameters iteratively. PSFNet is designed to fuse the image features in a progressive fashion using a stack of blocks having the same structure named the {\em progressive and selective feature block} (PSFB). A single PSFB updates the image features by a small amount, and a series of PSFBs updates them gradually, completing their fusion. 

For (ii), we design the PSFB that effectively performs the two operations playing the central roles in the feature fusion, i.e., {\em comparison} and {\em selection}. The former compares the input LDR images to identify their differences. The latter selects the images/regions based on the comparison results. These two operations are the key to successful HDR image computation since it is fundamentally correctly selecting images/regions that are well-exposed and properly eliminating inter-image misalignment and possible occlusions. 

The experiments demonstrate that the above approach can successfully produce ghosting-free HDR images. Our method can achieve better performance in terms of quantitative and qualitative evaluation than the popular algorithms on the commonly used public test datasets.

\begin{figure*}[h]
  \centering
  \includegraphics[width=\linewidth]{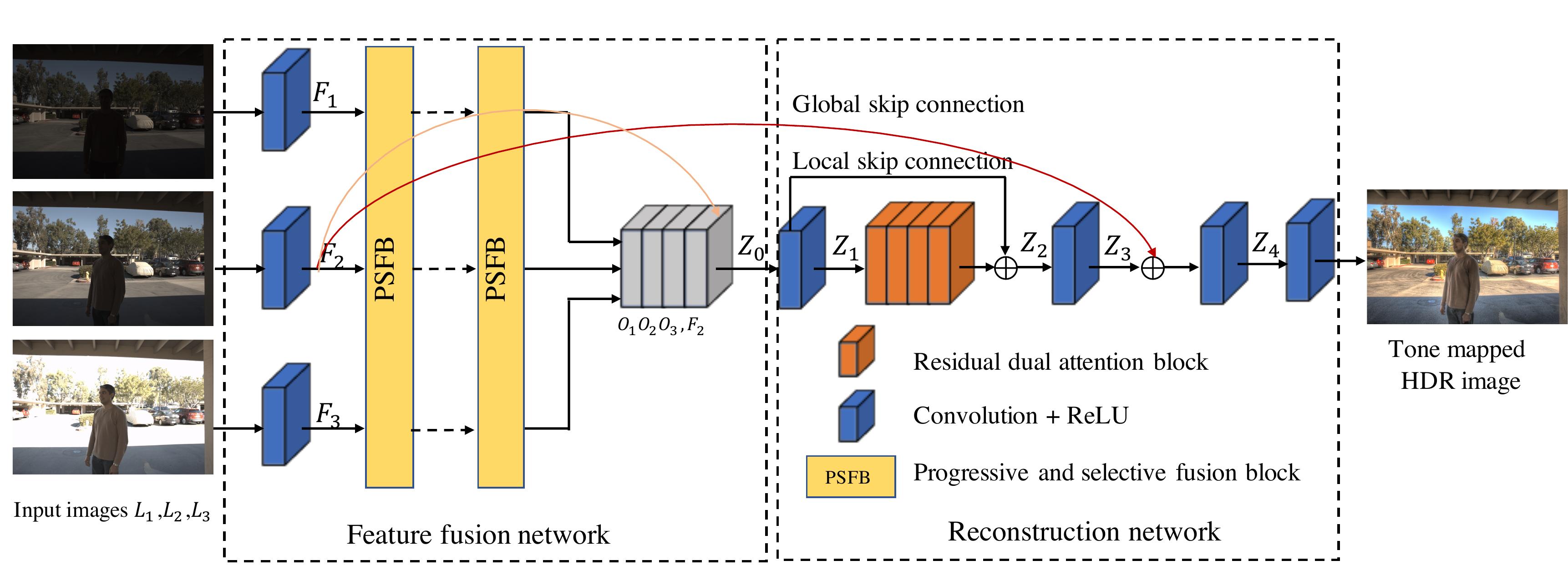}
  \caption{Archtecture of proposed network.}
 \label{Fig: arch}
\end{figure*}

\section{Related Work}
\subsection{Motion Detection and Removal}

Some methods classify all the pixels of the input LDR images as static or moving depending on whether they correspond to static background or moving foreground. They then merge the pixels found to be static while removing those found to be moving. In \cite{khan2006ghost}, weights are computed iteratively and then applied to pixels to determine their contribution to the final image. Heo et al. \cite{heo2010ghost} propose a method that assigns a weight to each pixel by computing a Gaussian-weighted distance to a reference pixel. Jacobs et al. \cite{jacobs2008automatic} propose a method that detects moving pixels by calculating the variance of different LDR images. Zhang et al. \cite{zhang2011gradient} propose to detect image motion by analyzing image gradients. Rank minimization \cite{lee2014ghost} and sparse representation \cite{yan2017high} have been employed to detect outliers, including moving pixels, and reconstruct the final HDR image. However, even when all pixels are classified correctly, removing the moving pixels makes it impossible to utilize all the information contained in the input images; the generated HDR images will inevitably lose some details. 

\subsection{Alignment based Methods}

These methods first align the input LDR images and then merge them to reconstruct an HDR image. The images are aligned either at the pixel level or the patch level. In \cite{bogoni2000extending}, the optical flow field is estimated and used for the alignment by warping the input images with them. In \cite{kang2003high}, a method for merging images is proposed to eliminate the artifacts of the alignment using optical flow. 
Sen et al. \cite{sen2012robust} formulate the HDR reconstruction as an energy-minimization task that jointly solves patch-based alignment and HDR image reconstruction. Hu et al. \cite{hu2013hdr} propose a method for image alignment using brightness and gradient information. Hafner et al. \cite{hafner2014simultaneous} propose an energy-minimization method that simultaneously computes the aligned HDR composite and accurate displacement maps. These methods tend to fail to deal with large motion and excessively dark or bright regions since the alignment process is vulnerable to them, generating artifacts in the aligned images. These methods employ a simple method for merging aligned LDR images, which cannot eliminate those artifacts. 

\subsection{Deep Learning based Methods}

As with other similar tasks, deep learning has been applied to HDR image generation. Eilertsen et al. \cite{eilertsen2017hdr} propose a deep network having the encoder-decoder structure for HDR image generation from a single image. It is proposed in \cite{endoSA2017} to synthesize multiple LDR images with different exposures from a single LDR image with a deep learning based method and use them to generate an HDR image. Such single-image-based methods tend to fail to reconstruct the textures on the saturated regions accurately. Kalantari et al. \cite{kalantari2017deep} propose using a convolutional neural network (CNN) for the task, which takes the LDR images for the input aligned in advance using optical flow. Wu et al. \cite{wu2018deep} propose to use a CNN with the U-net structure to reconstruct a ghosting-free HDR image without explicit alignment of the input images. Yan et al. \cite{yan2020deep} propose a non-local structure into a U-net to improve the accuracy of HDR image generation. Yan et al. \cite{yan2019attention} propose attention modules for improving the merging of input images with a reference image. Pu et al. \cite{pu2020robust} use deformable convolution \cite{zhu2019deformable}  across multiscale features to perform pyramidal alignment of input images and also use attention mechanism to reconstruct an HDR image from the aligned feature accurately.

\section{Proposed Method}

\subsection{Outline of the PSFNet }

Given a set of LDR images, $L_1, L_2,..., L_k$, of a dynamic scene with different exposures, the goal of HDR imaging is to reconstruct an HDR image $H$ aligned to a selected reference image $L_r$.
Following the settings in \cite{kalantari2017deep}, we consider the case of using three LDR images, $L_1$, $L_2$, and $L_3$, sorted in the order of exposures. We select $L_2$
as the reference image. 
Following \cite{kalantari2017deep}, before feeding them into the network, we map the LDR images into an HDR domain using gamma correction. 
To be speicfic, we map $L_i$ to $H_i$ as
\begin{equation}
    H_{i} = L_{i}^{\gamma}/t_i, \;\;
    i=1, 2, 3,
\end{equation}
where $\gamma=2.2$ \cite{poynton2012digital} and  $t_i$ is the exposure time of $L_i$.  
We then concatenate $L_i$ and $H_i$ in the channel dimension to get a six channel tensor $X_i=[L_i, H_i]$ for each of $i=1,2,3$ and input $X_1$, $X_2$, and $X_3$ to our network. We expect that $L_i$'s help identify image noises and/or saturated regions, whereas $H_i$'s help identify the differences from the reference image. 


Our network consists of two sub-networks, {\em the feature extraction network} and {\em the reconstruction network}, as shown in Figure~\ref{Fig: arch}. 
While the overall construction is similar to the encoder-decoder design employed in previous studies \cite{yan2019attention, pu2020robust}, we design the sub-networks with clearer intentions. Specifically, the feature fusion network fuses the input LDR images in their feature space, aiming to eliminate their misalignment and possible occlusions due to moving objects. The merging network reconstructs an HDR image from the fused feature. We will explain their details in what follows.


\subsection{Feature Fusion Network}

\begin{figure}[t]
  \centering
  \includegraphics[width=\linewidth]{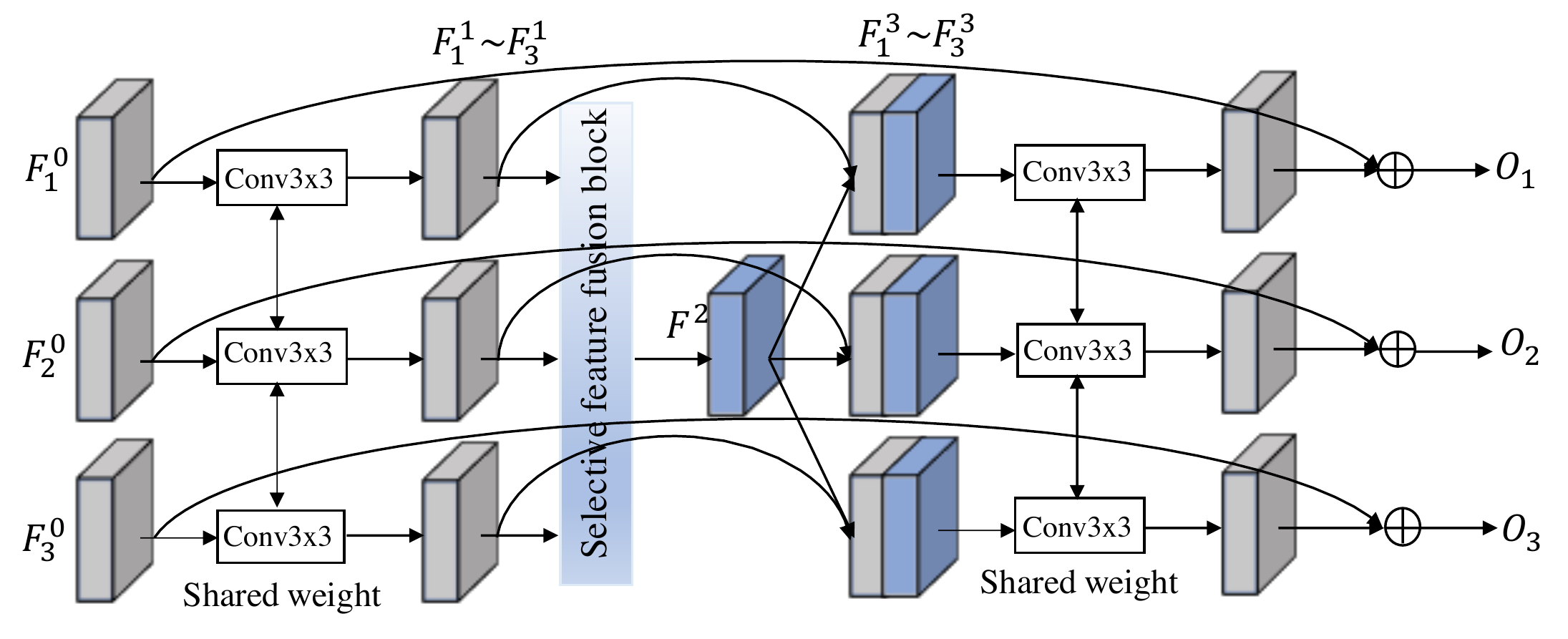}
\caption{Progressive and selective fusion block (PSFB). }
 \label{Fig: PFSB}
\end{figure}

The input LDR images are geometrically unaligned and could contain occlusions, which need to be eliminated. Previous studies leave it to the fusion of the input image features in the encoder part of a network, which is also intended to generate an HDR image from multiple input images. Most of them perform the feature fusion in a single step using a simple operation such as summation and concatenation. We think this leads to suboptimal results causing ghosting artifacts etc. For better feature fusion, we employ i) a multi-step approach that progressively fuses features and ii) a more suitable mechanism for the feature fusion. We explain the two below.

\subsubsection{Progressive Feature Fusion}

Initially extracting features from the input LDR images, the feature fusion network fuses their features in a progressive fashion using a stack of blocks having the same structure, as shown in Figure~\ref{Fig: arch}. We name the block {\em the progressive and selective fusion block} (PSFB). In our experiments, we use a stack of six PSFBs. 

Our intention behind this design is to split the difficult task of feature fusion into multiple steps, by which we attempt to make it easier for the network to learn to perform the task. It is  analogous to nonlinear optimization algorithms that update parameters iteratively. Previous studies of other tasks employed this idea of designing an architecture to gradually update estimates to get better results, e.g., RAFT (recurrent all-pairs field transforms) for optical flow estimation \cite{teed2020raft}. Our approach shares the same motivation. 

\subsubsection{Progressive and Selective Feature Block} 

Roughly speaking, the computation of an HDR image requires to conduct the following two: the selection of images/regions that are well-exposed (i.e., neither under nor over-exposed) and the elimination of inter-image misalignment and possible occlusions. For the latter, it will be necessary first to identify differences among the images. 

These computations will essentially reduce to two fundamental operations, {\em comparison} and {\em selection}, i.e., comparing the images to identify their differences and selecting them based on the comparison results. 
We design the PSFB to perform these two operations effectively; see Figure~\ref{Fig: PFSB}.

A PSFB performs the inter-image comparison in its second half. Specifically, we concatenate the fused feature computed in its first half with the individual image features and then apply convolution to each of the concatenated features, as shown in Figure~\ref{Fig: PFSB}. This computation will perform the above comparison since the fused feature should contain all the image information; the convolution will learn to compare the input image features and identify their differences.

\begin{figure}[t]
  \centering
  \includegraphics[width=\linewidth]{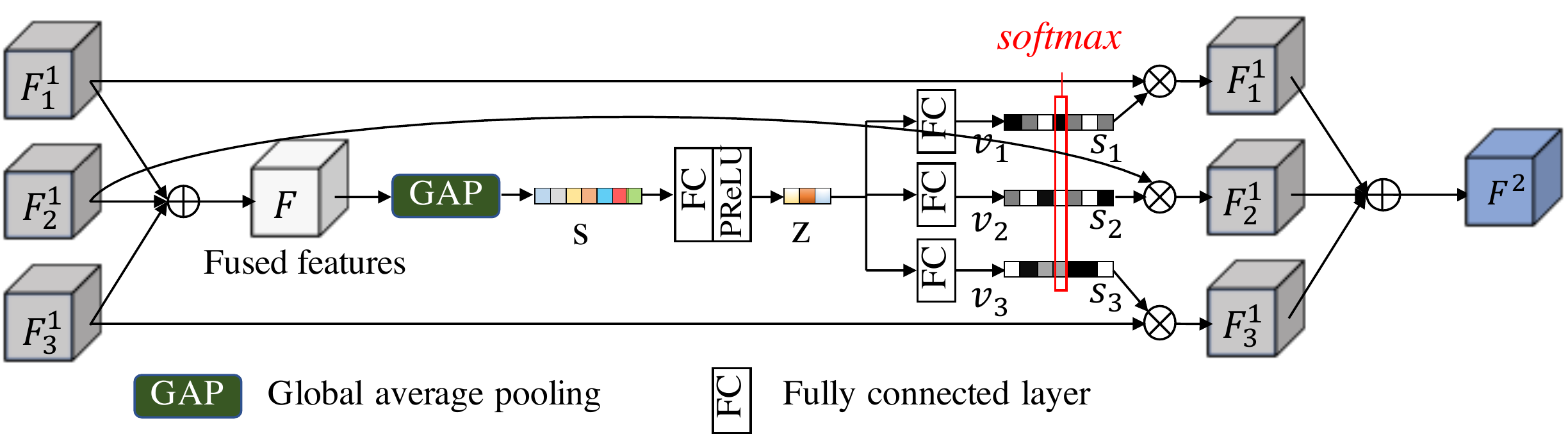}
  \caption{Selective feature fusion block (SFFB). 
  }
 \label{Fig: SKB}
\end{figure}

The PSFB performs the second operation of {\em selection} in a component named the {\em selective feature fusion block} (SFFB) in its first half. To design the SFFB, we borrow the feature fusion mechanism of the selective kernel convolution \cite{li2019selective}, which was developed to adaptively choose the size of convolution kernels (e.g., $3\times 3$ or $5\times 5$) in a convolution layer. Figure \ref{Fig: SKB} shows the design of SFFB. 

Previous studies fuse multiple features with simple methods such as summation and concatenation, which we think limits the expressive power of the network. Following \cite{li2019selective}, SFFB performs dynamic fusion via two operations, i.e., {\em fuse} and {\em select}. The fusion operator aggregates the input features by their summation followed by global average pooling and two fully-connected layers, yielding attention weights on the channels of the individual input features. The selection operator applies these attention weights to the input features. The attended features are added to form a fused feature, which is the final output of a SFFB.

It should be noted that while the order of comparison and selection is exchanged within a PSFB, the fused feature is necessary for the comparison of the image features as above, and their intra-block order does not matter since we stack multiple PSFBs as mentioned above.




\subsubsection{Details of Computation in PSFB}

Taking the first block as an example, we explain the detailed design of a PSFB here. Firstly, setting the input feature $F_{i}^0=F_i$ (as this is the first block), it updates $F_{i}^0$ into $F_{i}^1$ using a convolution layer with kernel size of $3\times 3$ as
\begin{equation}
    F_{i}^1 = Conv1(F_{i}^0).
\end{equation}
We then fuse features $\{ F_{1}^1, F_{2}^1, F_{3}^1\}$ using the above SFFB, yielding a fused feature $F^2$ as
\begin{equation}
    F^2 = {\color{black} SFFB}
    ([F_{1}^1, F_{2}^1, F_{3}^1]).
\end{equation}

We concatenate the fused feature $F^2$ with $F_i^1$ as $F_i^3=[F^2,F_i^1]$ $(i=1,2,3)$.
Thus, $F_{i}^3$ contains the individual image feature and the fused feature. 
We use a convolution layer with the kernel size of $3\times 3$ to convert it to an feature $O_i$
as 
\begin{equation}
    O_{i} = Conv2(F_{i}^3)+F_{i}^0
\end{equation}
We use “$+F_{i}^0$” to represent a 
residual connection
\cite{he2016deep};
the output of $Conv2()$ has the same size as $F_i^0$. 

\subsubsection{Summary of the Feature Fusion Network}


As shown in Figure~\ref{Fig: arch}, the input to the feature fusion network are
the three 6-channel input images $X_1$, $X_2$, and $X_3$ corresponding to the input LDR images. The network first extracts a feature map with $N=64$ channels from each using a shared convolutional layer, yielding $F_1$, $F_2$, and $F_3$. These are inputted to the stack of six PSFBs. As explained above, a single PSFB updates the image features by a small amount, and thus the features are progressively fused in the PSFB stack. The input and output of all the PSFBs have the same size and channels. The features outputted from the last PSFB are concatenated as $Z_0=[O_1, O_2, O_3, F_2]$ and then inputted to the reconstruction network.

\subsection{Reconstruction Network}
In \cite{yan2019attention}, three dilated residual dense blocks are used to decode the encoded image feature, reconstructing an HDR image. Although their method achieves good performance, it tends to consume a large amount memory,
especially when the input image size is large. To cope with this, we employ a residual block with dual attention mechanism \cite{zamir2020cycleisp} consisting of channel attention \cite{hu2018squeeze} and spatial attention \cite{woo2018cbam}. The residual dual attention mechanism is known to work well for superresolution \cite{woo2018cbam} and denoising \cite{zamir2020cycleisp}. We expect it to work for our case because of their similarity. 

\begin{figure}[tbh]
  \centering
  \includegraphics[width=\linewidth]{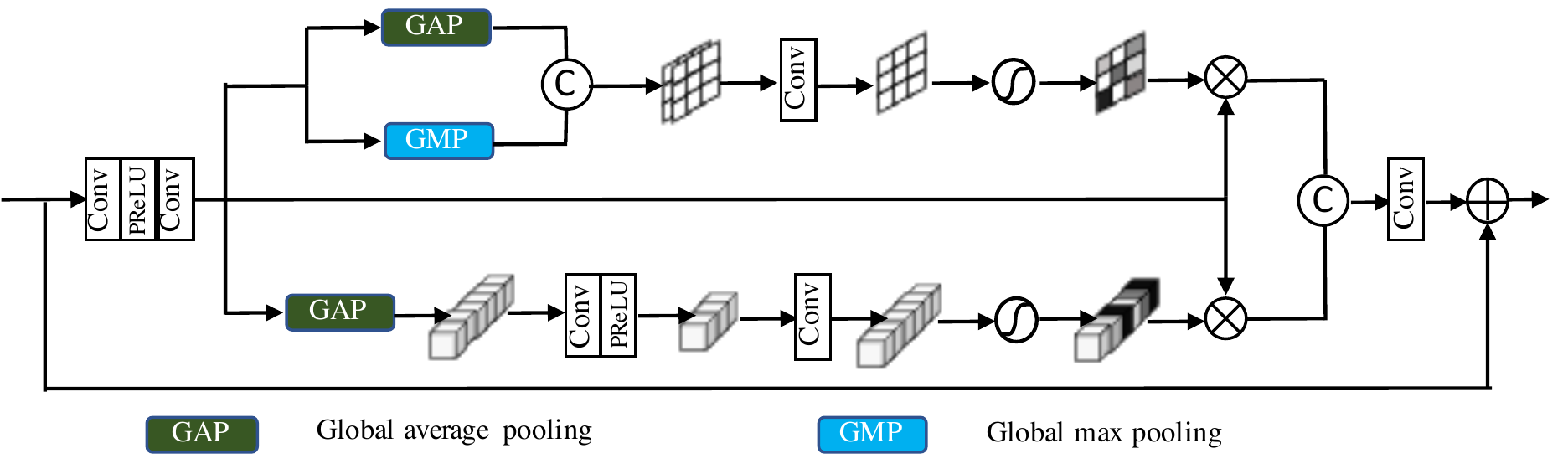}
  \caption{Dual attention block}
 \label{Fig: DAB}
\end{figure}

\begin{figure}[tbh]
  \centering
  \includegraphics[width=\linewidth]{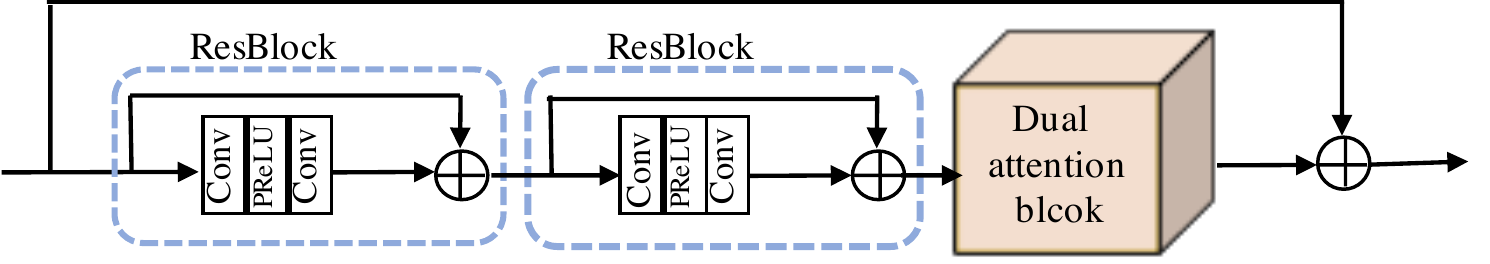}
  \caption{Residual dual attention block}
 \label{Fig: RDAB}
\end{figure}

As shown in Figure~\ref{Fig: arch}, the reconstruction network takes the concatenated features $Z_{0}=[O_1, O_2, O_3, F_2]$.
It consists of a series of a convolution layer, four residual dual attention blocks, and three additional convolution layers with a local and a global skip connections. The structure of a dual attention block is shown in Figure~\ref{Fig: DAB} and the structure of a residual dual attention block is shown in Figure~\ref{Fig: RDAB}. Applying a sigmoid function to the output of the last convolution layer, the reconstruction network outputs an HDR image. 




\subsection{Loss Function}

Following \cite{kalantari2017deep}, we consider the optimization in the domain of tone-mapped HDR images. It produces better results with fewer artifacts in the dark regions than the optimization in the original domain of HDR images.
To be specific, we employ $\mu$-law for tone mapping loss, which is formulated as,
\begin{equation}
    T(H) = \frac{log(1+\mu H)}{log(1+\mu)} ,
\end{equation}
where $\mu$ is set to $5,000$ throughout our experiments. It is also reported in \cite{yan2019attention} that minimizing the $L_1$ norm between the predicted HDR image $\hat{H}$ and its ground truth $H$ in the tone-mapped domain works better than others.
Following their study, we employ the following loss 
\begin{equation}
    L=\lVert T(\hat{H})-T(H) \rVert_1.
\end{equation}

\section{Experiments}

\subsection{Experimental Settings}

\subsubsection{Training data} We train our network on the dataset of Kalantari and Ramamoorthi \cite{kalantari2017deep}, which consists of indoor and outdoor scenes. It includes 74 samples for training and 15 samples for testing. We use the former for training the PSFNet. Each sample includes ground truth HDR images and three LDR images with exposure biases of $\{-2, 0, +2\}$ or $\{-3, 0, +3\}$. Following the standard procedure of recent studies \cite{kalantari2017deep,yan2019attention,pu2020robust}, we resize all the images to $1,000\times 1,500$ pixels.

\begin{table*}[tb]
  \caption{Comparison of the methods on the test set of \cite{kalantari2017deep}. The primary metrics are PSNR-$\mu$, SSIM-$\mu$, and HDR-VDP-2; see Sec.~\ref{sec:metrics} for more detailss. }
  \begin{tabular}{cccccc}
    \toprule
Methods & PSNR-$\mu$ & {\color{gray} PSNR-L}  & SSIM-$\mu$ & {\color{gray} SSIM-L} &HDR-VDP-2\\
\midrule
TMO \cite{endoSA2017} &  8.3120 & {\color{gray} 8.8459}  &  0.5029 & {\color{gray} 0.0924} & 44.3345\\
HDRCNN \cite{eilertsen2017hdr} &  13.7054 & {\color{gray} 13.8956} & 0.5924 & {\color{gray} 0.3456} & 47.5690\\

Sen \cite{sen2012robust} & 40.9689 & {\color{gray} 38.3425} & 0.9859  & {\color{gray} 0.9764} & 60.3463 \\
Kalantari \cite{kalantari2017deep}& 42.7177 & {\color{gray} 41.2200} & 0.9889 & {\color{gray}  0.9829} & 61.3139 \\
Wu \cite{wu2018deep}   &  41.9977  & {\color{gray} \textbf{41.6593} }& 0.9878 & {\color{gray} 0.9860} & 61.7981 \\

AHDR \cite{yan2019attention} & 43.7013 & {\color{gray}41.1782 }&  0.9905 & {\color{gray} 0.9857} & 62.0521 \\
PAN \cite{pu2020robust}      & 43.8487 & {\color{gray}41.6452 }&  0.9906 & {\color{gray} \textbf{0.9870}} & 62.5495 \\
PSFNet          & \textbf{44.0613} & {\color{gray} 41.5736} &  \textbf{0.9907} & {\color{gray} 0.9867 }& \textbf{63.1550} \\ 

  \bottomrule
\end{tabular}
\label{Table: main}

\end{table*}

\subsubsection{Testing data}  Following recent studies, we choose the datasets for testing. We evaluate methods on the 15 scenes of the dataset of \cite{kalantari2017deep}; we conduct quantitative evaluation using the provided ground truths. We also use the datasets of Sen et al. \cite{sen2012robust} and Tursun et al. \cite{tursun2016objective}, which do not contain ground truths; we use them for qualitative evaluation.

\subsubsection{Evaluation metrics}  
\label{sec:metrics}

It is argued in \cite{kalantari2017deep} that HDR image generation methods should be evaluated in the tone-mapped domain, as we usually use generated HDR images after tone-mapping. Following this argument, we use PSNR-$\mu$ and SSIM-$\mu$ for primary metrics, which are PSNR and SSIM values in the tone-mapped domain. We show PSNR and SSIM in the linear domain, denoted by PSNR-L and SSIM-L, for completeness. We also show HDR-VDP-2 \cite{mantiuk2011hdr}, which is designed to evaluate the quality of HDR images.

It should be noted that there is a limitation in the evaluation based on PSNR-$\mu$ etc. The recent studies, including ours, aim at adequately dealing with ghosting artifacts. However, the artifacts usually appear in a small area of an image, and they often have only small impacts on these metrics. HDR-VDP-2 may better evaluate the image quality in that case. To supplement the quantitative evaluation, we also show the results of qualitative comparisons.


\subsection{Implementation Details}

The training data are first cropped into patches of $256\times256$ pixel size with stride of 128 pixels. We employ rotation and flipping for data augmentation to avoid over-fitting. We use the Adam optimizer \cite{KingmaB14} with $\beta_1=0.9$ , $\beta_2=0.999$, initial learning rate $=10^{-4}$ and set the batch size to 8. We
perform training for 210 epochs 
and employ the cosine annealing strategy \cite{loshchilov2016sgdr} to steadily
decrease the learning rate from initial value to $1\times10^{-6}$.
We conducted all the experiments using PyTorch \cite{paszke2017automatic} on NVIDIA GeForce RTX 2080 GPUs.

\subsection{Comparison with the State-of-the-art Methods}

We evaluate the proposed method and compare it with
previous methods including the state-of-the-art. We use Kalantari’s testset \cite{kalantari2017deep} for quantitative evaluation and the above datasets without ground truths \cite{sen2012robust,tursun2016objective} for qualitative evaluation. The compared methods are as follows: two single image HDR imaging methods, TMO \cite{endoSA2017} and HDRCNN \cite{eilertsen2017hdr}, and five multi-image HDR imaging methods, the patch-based method \cite{sen2012robust}, the flow-based method with CNN merger \cite{kalantari2017deep}, the U-net structure without optic flow \cite{wu2018deep}, the attention-guide method (AHDR) \cite{yan2019attention}, and pyramidal alignment network (PAN) \cite{pu2020robust}. For all methods, we used the authors' code for testing comparison, except \cite{pu2020robust} since their code is not available as of the time of writing this paper.

\subsubsection{Evaluation on Kalantari et al.’s Dataset}
\begin{figure}[tb]
  \centering
  \includegraphics[width=\linewidth]{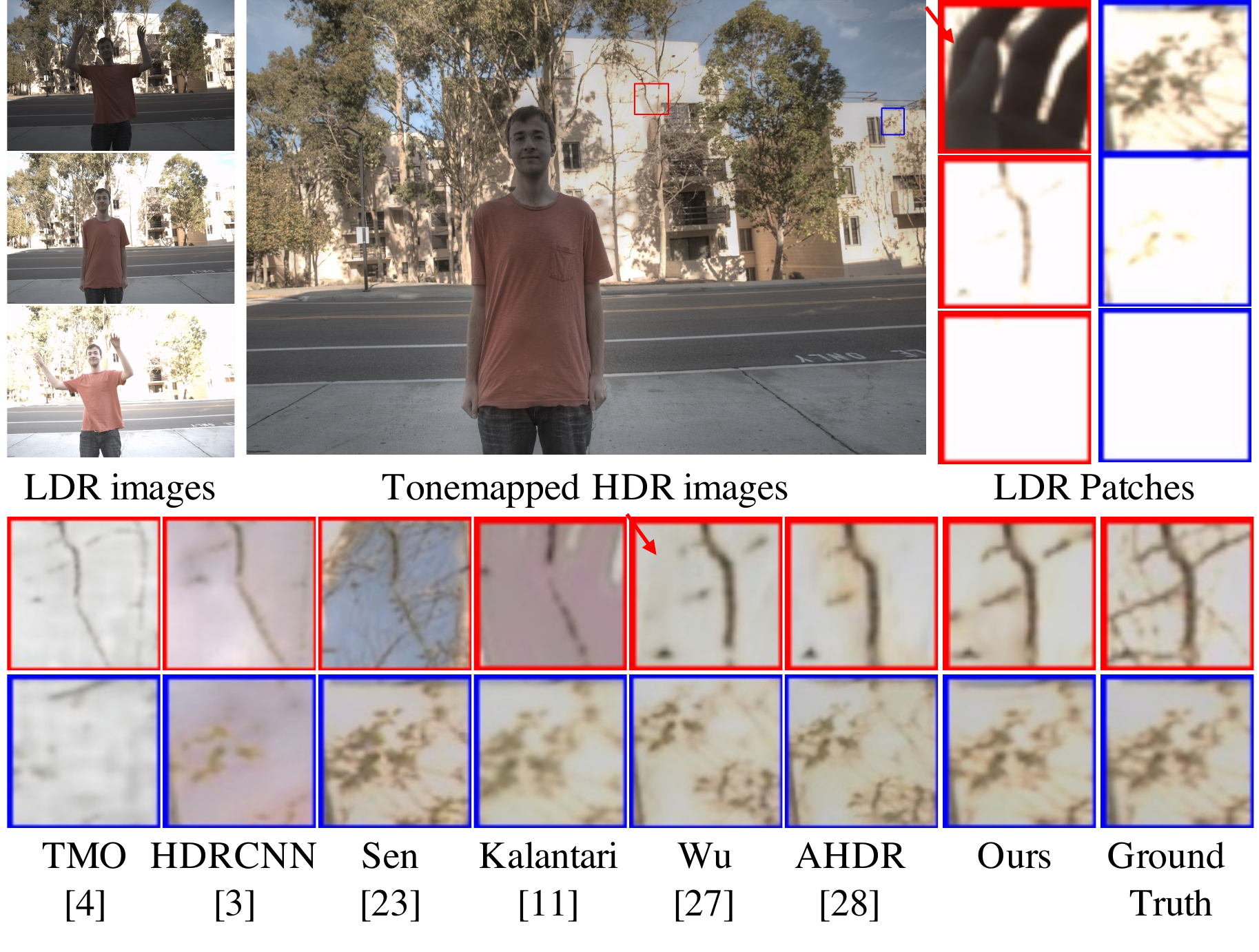}
  \vspace*{-6mm}
  \caption{Results for ``Building'' from the test set of \cite{kalantari2017deep}. Upper row from left to right: the three input LDR images, the HDR image produced by the proposed method, and (zoomed-in) LDR image patches with two identical positions/sizes (in red and blue). Lower row: the same patches of the HDR images produced by different methods.}
 \label{Fig: 1}
\end{figure}

Table \ref{Table: main} shows the quantitative evaluation on the test set of \cite{kalantari2017deep}, i.e., averaged values over 15 test scenes. It is seen that the proposed method achieves better performance than others in the primary metrics, PSNR-$\mu$, SSIM-$\mu$, and HDR-VDP-2. Figure \ref{Fig: 1} shows an example of qualitative comparisons. The input LDR images contain saturated background and foreground motions. It is observed from the results of the single-image methods, TMO \cite{endoSA2017} and HDRCNN \cite{eilertsen2017hdr}, which uses the reference image alone, that while they can avoid the ghosting artifacts, they cannot recover detailed textures; they also suffer from color distortion. The patch based method of Sen et al. \cite{sen2012robust} fails to find correct patches, generating some artifacts. The method of Kalantari et al. \cite{kalantari2017deep} cannot completely eliminate the effects of the occlusion. The method of Wu et al. \cite{wu2018deep} and AHDR \cite{yan2019attention} produce better results but fail to recover the fine details of the texture. Our method produces the best results; it produces less color distortion and recovers the textures more accurately.

\subsubsection{Evaluation on Datasets w/o Ground Truth}

\begin{figure}[tb]
  \centering
  \includegraphics[width=\linewidth]{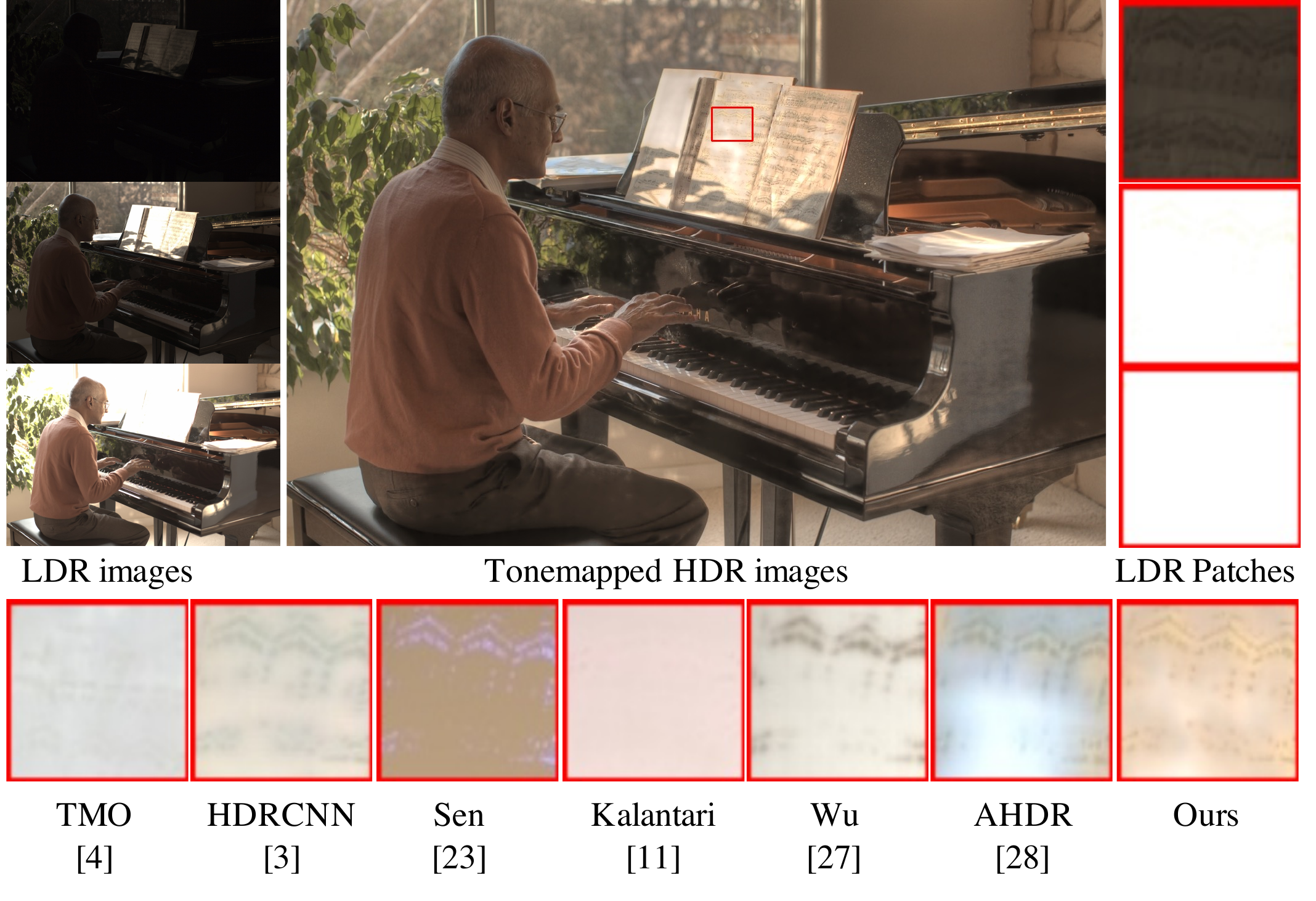}
  \vspace*{-6mm}
  \caption{Results for ``PianoMan'' from the dataset of \cite{sen2012robust}. See Figure~\ref{Fig: 1} for the explanation of the panels.}
 \label{Fig: 2}
\end{figure}

\begin{figure}[tb]
  \centering
  \includegraphics[width=\linewidth]{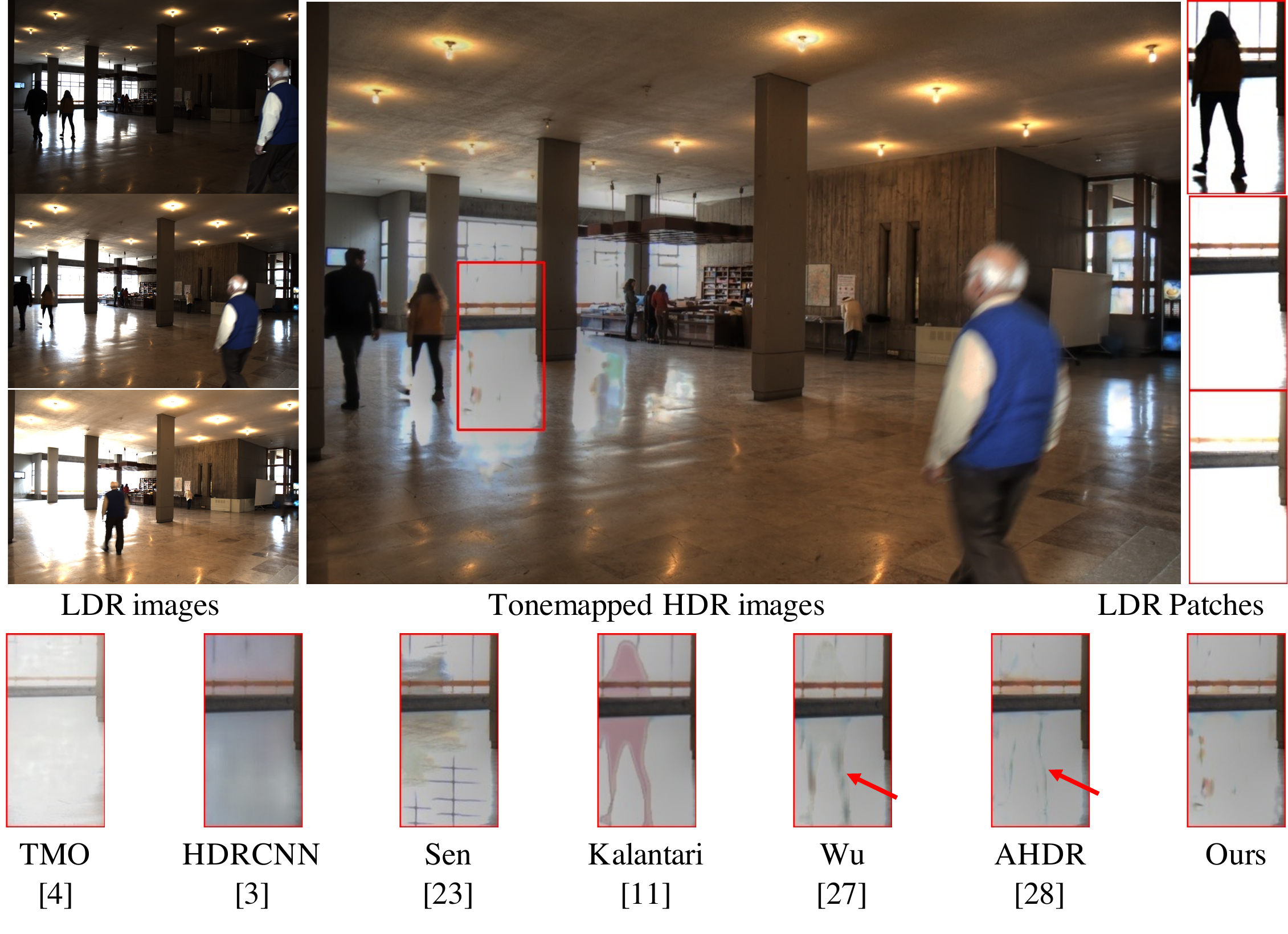}
  \vspace*{-6mm}
  \caption{Results for an image from the dataset of \cite{tursun2016objective}. See Figure~\ref{Fig: 1} for the explanation of the panels.}
 \label{Fig: 3}
\end{figure}

We also show qualitative comparisons using Sen’s \cite{sen2012robust} and Tursun’s \cite{tursun2016objective} datasets, which do not have ground truths. The results are shown in Figure \ref{Fig: 2} and Figure \ref{Fig: 3}. The single image methods, TMO \cite{endoSA2017} and HDRCNN \cite{eilertsen2017hdr}, fail to recover a sharp image and suffer from color distortion. The patch based method (Sen et al \cite{sen2012robust}) produces severe artifacts in the saturated area and ghosting artifacts in the areas undergoing large motion. The same is true for the method of Kalantari et al. \cite{kalantari2017deep}; it produces artifacts in the areas undergoing large motion and fails to recover the details of the saturated areas. These are arguably because of the possible misalignment of optical flows and the limitation of the merging method. The results of Wu et al.’s method \cite{wu2018deep} tend to show over-smoothness and yield ghosting artifacts on the large motion areas. AHDR \cite{yan2019attention} yields artifacts in the saturated areas and also suffers from ghosting artifacts due to large motion. On the other hand, our method produces good results with noticeably reduced geometric and color distortions compared with others.

\subsection{Ablation Study}

We conducted experiments to evaluate the components of PSFNet. 
First, we compare different methods for feature fusion in the PSFB. To be specific, we replace the SFFB with concatenation or summation and evaluate the performance. Table \ref{Table: fusion} shows the results. It is observed that the SFFB yields better PSNR values, although there is little difference in SSIM values.
\begin{table}[t]
  \caption{Comparison of fusion methods on the test set of \cite{kalantari2017deep}.}
  \label{Table: fusion}
  \begin{tabular}{ccccc}
    \toprule
Alignment Methods &  PSNR-$\mu$ & PSNR-L  &SSIM-$\mu$ & SSIM-L  \\
    \midrule
Summation        & 43.9789 & 41.4092 & 0.9904 & 0.9866 \\   
Concatenation    & 43.9560 & 41.4579 & 0.9907 & 0.9867   \\
SFFB             & 44.0613 & 41.5736 & 0.9907 & 0.9867   \\ 
  \bottomrule
\end{tabular}
\end{table}

\begin{figure}[tb]
  \centering
  \includegraphics[width=\linewidth]{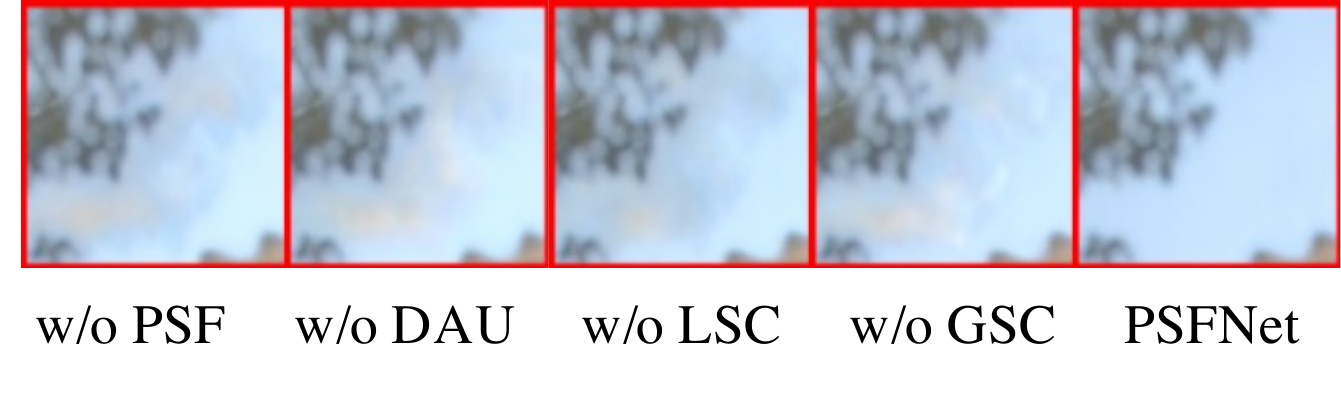}
  \vspace*{-6mm}
  \caption{Results obtained by ablated networks.}
 \label{Fig: 4}
\end{figure}

We also evaluate individual components in PSFNet. We eliminate either the stack of PSFBs, the local skip connection (LCS), or the global skip connection (GSC) from the feature fusion network. When we eliminate the PSFB stack, we use the feature maps $F_i$'s instead of $O_i$'s. We also ablate the DAB from the reconstruction network. Table \ref{Table: diff_component} shows the results. It is seen that PSFBs and DAB are essential to achieve the best performance and the skip connections (LSC and GSC) show modest contributions. Figure \ref{Fig: 4} shows examples of zoomed-in patches of an HDR image produced by the ablated networks. It is seen that color distortion emerges except the PSFNet (with full components).

\begin{table}[tb]
  \caption{Ablation study using the test set of \cite{kalantari2017deep}.}
 \label{Table: diff_component}
  \begin{tabular}{ccccc}
    \toprule
Methods  & PSNR-$\mu$ & PSNR-L & SSIM-$\mu$ & SSIM-L \\
\midrule
w/o PSFBs  & 43.3586 & 40.9648 & 0.9902 & 0.9864 \\ 
w/o DAB  & 43.7917 & 41.0672 & 0.9903 & 0.9856  \\ 
w/o LSC  & 43.9923 & 41.4289 & 0.9906 & 0.9864 \\ 
w/o GSC  & 44.0125 & 41.3756 & 0.9907 & 0.9866 \\ 
PSFNet   & 44.0613 & 41.5736 & 0.9907 & 0.9867 \\ 

  \bottomrule
\end{tabular}

\end{table}



Finally, we examine how the number of PSABs in the feature fusion network affects the performance. Table \ref{Table: number_blocks} shows the results. It is seen that there is a peak around 6 and 7 blocks. We conclude that stacking multiple PSABs does contribute to better performance and too large a number of blocks does not lead to a good result. 


\begin{table}[tb]
  \caption{Number of PSFBs used in the feature fusion network.}
 \label{Table: number_blocks}

  \begin{tabular}{ccccc}
    \toprule
Number of Blocks  & PSNR-$\mu$ & PSNR-L  & SSIM-$\mu$ & SSIM-L \\
    \midrule
4       & 43.9610 & 41.1000 & 0.9907 & 0.9865 \\
5       & 43.9895 & 41.3123 & 0.9907 & 0.9865 \\   
6       & 44.0613 & 41.5736 & 0.9907 & 0.9867  \\ 
7       & 44.0504 & 41.5091 & 0.9906 & 0.9867\\
8       & 43.9737 & 41.3107 & 0.9906 & 0.9860\\
  \bottomrule
\end{tabular}

\end{table}

\begin{figure}[tb]
  \centering
  \includegraphics[width=\linewidth]{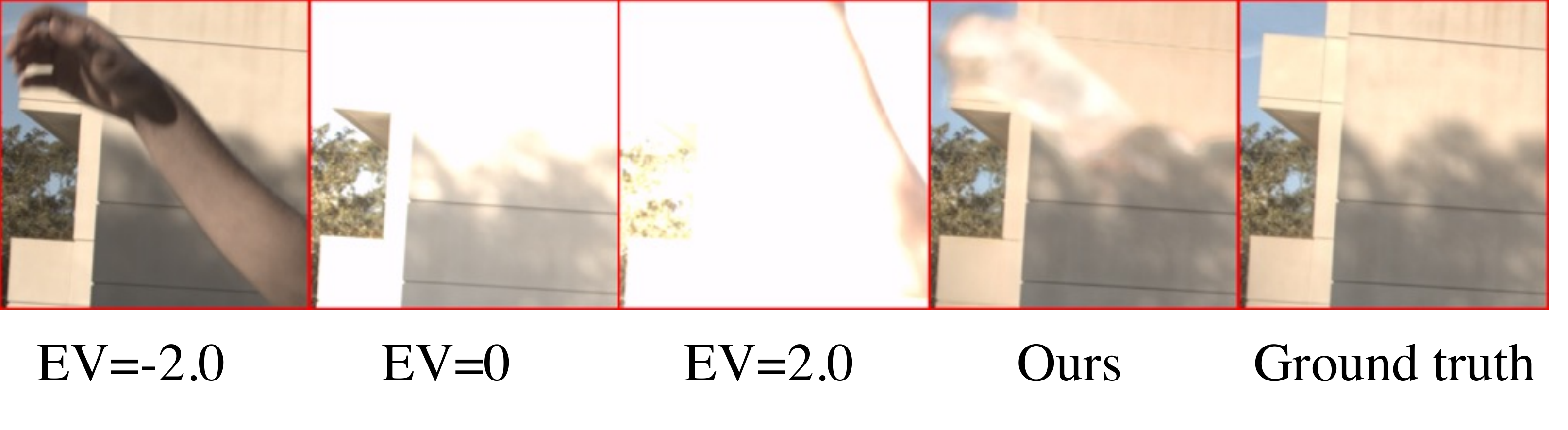}
  \vspace*{-6mm}
  \caption{Failure case of proposed method.}
 \label{Fig: limitation}
\end{figure}
\subsection{Limitation}

Although our method achieves good performance, as reported above, there are several cases that it cannot handle well. Figure \ref{Fig: limitation} shows an example, where the generated image contains artifacts on the region occluded by the hand in one of the input images. The artifacts emerge because the other images do not provide useful information about the occluded area due to overexposure. Our method does not work well for such cases; they might be better formulated as image inpainting. More visualization results can be found at \url{https://github.com/yarqian/PSFNet/}.

\section{Summary and Conclusion}

In this paper, we have proposed a new method for generating an HDR image of a dynamic scene from its LDR images. When employing deep learning, the problem reduces to first fusing the features of the input images and then reconstructing an HDR image from the fused feature. The first step of feature fusion plays a central role in generating good quality HDR images. Considering the complexity and difficulty with it, we proposed a network named PSFNet that gradually fuses the image features in multiple steps with a stack of computational blocks and the design of the component block that can effectively perform the two operations fundamental to the HDR image generation, i.e., comparing and selecting appropriate images/regions. The experimental results have validated the effectiveness of the proposed approach. 


\begin{acks}
  This work was partly supported by JSPS KAKENHI Grant Number 20H05952 and JP19H01110.
\end{acks}


\bibliographystyle{ACM-Reference-Format}
\balance
\bibliography{acmart}




\end{document}